\documentclass[runningheads]{llncs}

\usepackage{eccv}

\usepackage{eccvabbrv}
\usepackage{graphicx}
\usepackage{booktabs}
\usepackage{multirow}
\usepackage{colortbl}
\usepackage[accsupp]{axessibility}
\usepackage{hyperref}
\usepackage{tcolorbox}
\usepackage{subcaption}
\usepackage{tabularx}
\usepackage{array}
\usepackage{svg}
\usepackage{enumitem}
\usepackage{amsmath}
\usepackage{algorithm}
\usepackage{algorithmic}
\usepackage{longtable}
\usepackage{float}
\usepackage{placeins}
\usepackage{xcolor}

\tcbuselibrary{skins,breakable}

\definecolor{casebg}{RGB}{248,248,248}
\definecolor{caseframe}{RGB}{150,150,150}
\definecolor{casetitlebg}{RGB}{110,110,110}
\definecolor{gttext}{RGB}{35,95,45}
\definecolor{predtext}{RGB}{145,35,35}
\definecolor{accentred}{RGB}{190,40,40}

\newcommand{\gtlabel}{\textbf{\textcolor{gttext}{Ground Truth}}}
\newcommand{\predlabel}{\textbf{\textcolor{predtext}{Model Prediction}}}
\newcommand{\gtcell}[1]{{\color{gttext} #1}}
\newcommand{\predcell}[1]{{\color{predtext} #1}}
\newtcolorbox{casebox}[1]{
  enhanced,
  breakable,
  colback=casebg,
  colframe=caseframe,
  colbacktitle=casetitlebg,
  coltitle=white,
  boxrule=0.6pt,
  arc=2pt,
  left=8pt,
  right=8pt,
  top=7pt,
  bottom=7pt,
  title={#1},
  fonttitle=\bfseries
}

\begin{document}

\title{HighlightBench: Benchmarking Markup-Driven Table Reasoning in Scientific Documents}
\titlerunning{HighlightBench}

\author{
Lexin Wang\inst{1},
Shenghua Liu\inst{1}\thanks{\fontsize{10}{12}\selectfont Corresponding author.},
Yiwei Wang\inst{2},
Yujun Cai\inst{3},\\[0.4em]
Yuyao Ge\inst{1},
Jiayu Yao\inst{1},
Jiafeng Guo\inst{1},
Xueqi Cheng\inst{1}
}
\authorrunning{L. Wang et al.}
\institute{
Institute of Computing Technology, Chinese Academy of Sciences \and
University of California, Merced \and
University of Queensland\\[0.4em]
\url{https://highlightbench.netlify.app/}
}

\maketitle
\pagestyle{plain}

\begin{abstract}
Visual markups such as highlights, underlines, and bold text are common in table-centric documents. Although multimodal large language models (MLLMs) have made substantial progress in document understanding, their ability to treat such cues as explicit logical directives remains under-explored. More importantly, existing evaluations cannot distinguish whether a model fails to see the markup or fails to reason with it. This creates a key blind spot in assessing markup-conditioned behavior over tables. To address this gap, we introduce HighlightBench, a diagnostic benchmark for markup-driven table understanding that decomposes evaluation into five task families: Markup Grounding, Constrained Retrieval, Local Relations, Aggregation \& Comparison, and Consistency \& Missingness. We further provide a reference pipeline that makes intermediate decisions explicit, enabling reproducible baselines and finer-grained attribution of errors along the perception-to-execution chain. Experiments show that even strong models remain unstable when visual cues must be consistently aligned with symbolic reasoning under structured output constraints.

\end{abstract}

\section{Introduction}
\label{sec:intro}

In academic papers, technical reports, and data analysis workflows, tables are often annotated with visual markups such as highlights, underlines, bold text, and arrows to emphasize key information. Reading such tables involves more than identifying values or parsing row--column structure: these markups are routinely used as functional cues that determine which entries should be retrieved, compared, filtered, or verified. For multimodal large language models (MLLMs) \cite{alayrac2022flamingo,dai2023instructblip,wang2024cogvlm}, this setting is fundamentally different from standard TableQA and broader multimodal capability evaluation \cite{pasupat2015compositional,zhong2017seq2sql,herzig2020tapas,liu2024mmbench}. Beyond recognizing table content, a model must correctly bind a queried visual cue to the intended structural unit, translate the cue into a constraint that is executable over table structure, and preserve this constraint throughout downstream reasoning.

A concrete gap emerges in practice: recent studies report that even strong models can produce plausible or correct answers without faithfully grounding the queried cue to the intended table region, especially when the cue is weak, ambiguous, or absent~\cite{zhou2025benchmarking}. In these cases, correctness does not certify cue-conditioned reasoning, and answer-only success can mask shortcut behaviors that bypass the cue. This mismatch motivates evaluations that disentangle cue perception from cue-conditioned execution, making markup-conditioned behavior measurable rather than implicit.

Recent multimodal table and document benchmarks have substantially improved realism and reasoning coverage \cite{zheng2024multimodal,zhu2025tableeval,wu2025tablebench,wang2024charxiv}. However, strong performance on these benchmarks still does not tell us whether a model truly uses embedded markups as reasoning constraints. It may instead arrive at the correct answer by exploiting dataset regularities, priors, or heuristics that are largely independent of the markup. This ambiguity is the core blind spot: answer-centric scores can conflate qualitatively different behaviors---failing to notice the markup, noticing it but mis-binding it to structure, or answering correctly while effectively ignoring it---and thus systematically misestimate capability in markup-sensitive settings.

To address this gap, we introduce \textbf{HighlightBench}, a diagnostic benchmark for markup-conditioned reasoning over tables. HighlightBench decomposes markup-conditioned table reasoning into five automatically scorable task families: Markup Grounding, Constrained Retrieval, Local Relations, Aggregation \& Comparison, and Consistency \& Missingness. This design separates grounding, constraint usage, and downstream execution, enabling fine-grained failure localization beyond end-to-end answer accuracy.

\begin{figure}[t]
    \centering
    \includegraphics[width=\columnwidth]{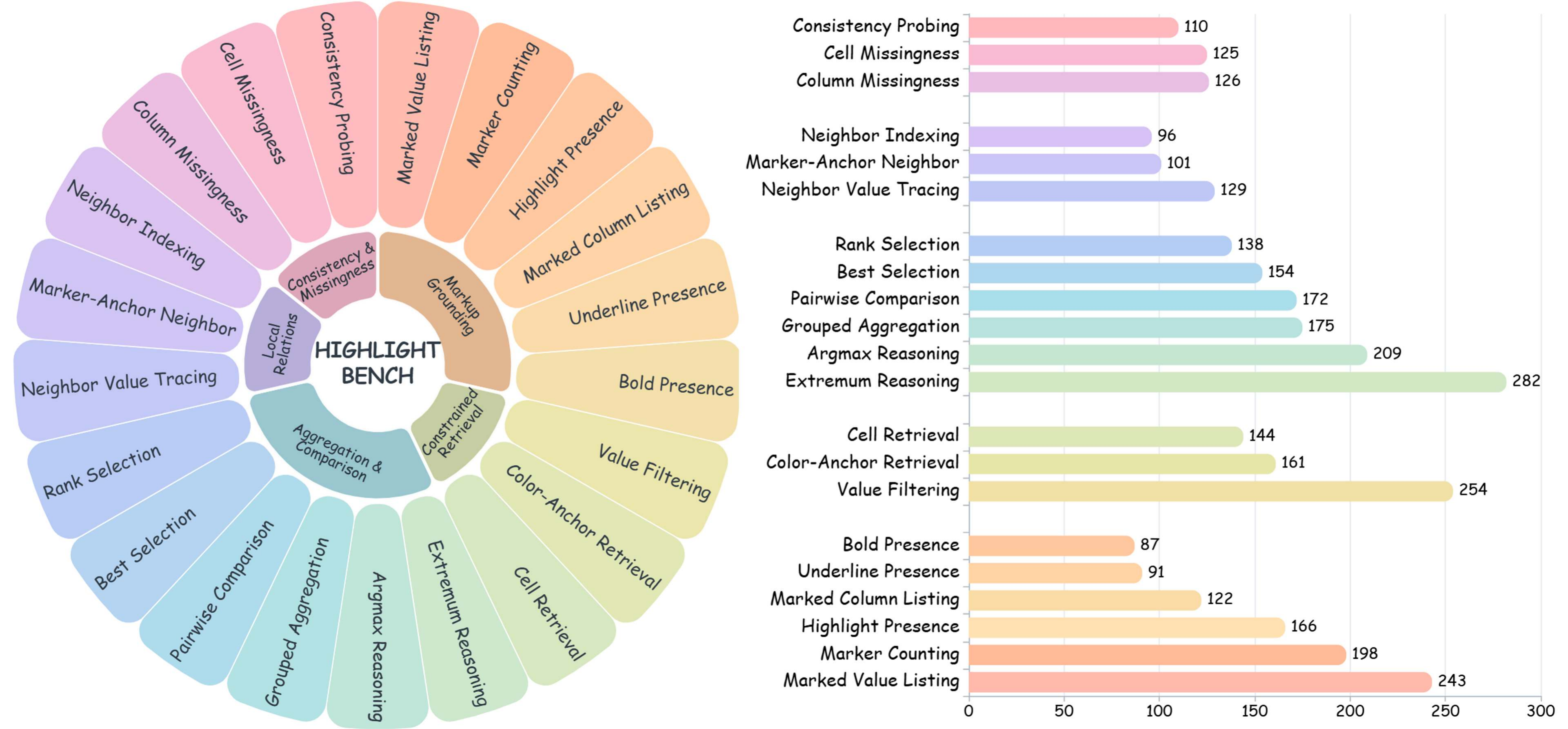}
    \caption{
    Overview of \textbf{HighlightBench}. The benchmark contains five complementary task families with question counts. This design gives transparent coverage of markup-conditioned reasoning.
    }
    \label{fig:highlightbench_overview}
\end{figure}

HighlightBench combines expert-collected real-world samples with controllable synthetic instances. The real-world subset contains markup-rich tables selected from scientific papers and paired with manually written questions, preserving natural markup patterns and realistic table irregularities. The synthetic subset is generated from structured templates, enabling controlled difficulty, reproducible settings, and counterfactual samples created by varying markup conditions under fixed table content.

We also provide a reference pipeline to support reproducible baselines and interpretable diagnostics. Given an input table image and its associated question, the pipeline produces a structured intermediate representation and a schema-constrained output, together with intermediate signals that expose where failures occur along the processing chain. This makes it possible to compare end-to-end model behavior with a more inspectable alternative and to attribute errors to specific stages.

Systematic evaluation on HighlightBench shows that current multimodal models still exhibit persistent weaknesses in markup alignment and constraint-aware reasoning. These weaknesses become especially visible when accurate grounding, correct computation, and format-compliant outputs are required simultaneously. In contrast, the reference pipeline improves stability on a subset of tasks and provides clearer intermediate error signals than end-to-end answers alone. Overall, HighlightBench turns markup-conditioned table reasoning into a scalable, automatically scorable, and diagnosable evaluation problem.

\paragraph{Contributions.}
\begin{itemize}[leftmargin=*, noitemsep]
    \item We formalize markup-conditioned reasoning over tables as a standalone evaluation target and introduce \textbf{HighlightBench}, a benchmark designed specifically for this setting.
    \item We construct a benchmark that combines expert-collected real-world samples with controllable synthetic instances, enabling automatically scorable evaluation, controlled difficulty, and finer-grained error attribution beyond end-to-end accuracy.
    \item We provide a reference pipeline that produces schema-constrained outputs with intermediate signals, serving both as a reproducible baseline and as a diagnostic interface for analyzing failure modes in current multimodal models.
\end{itemize}

\section{Related Work}
\label{sec:related}

\begin{figure*}[t]
    \centering
    \includegraphics[width=\textwidth]{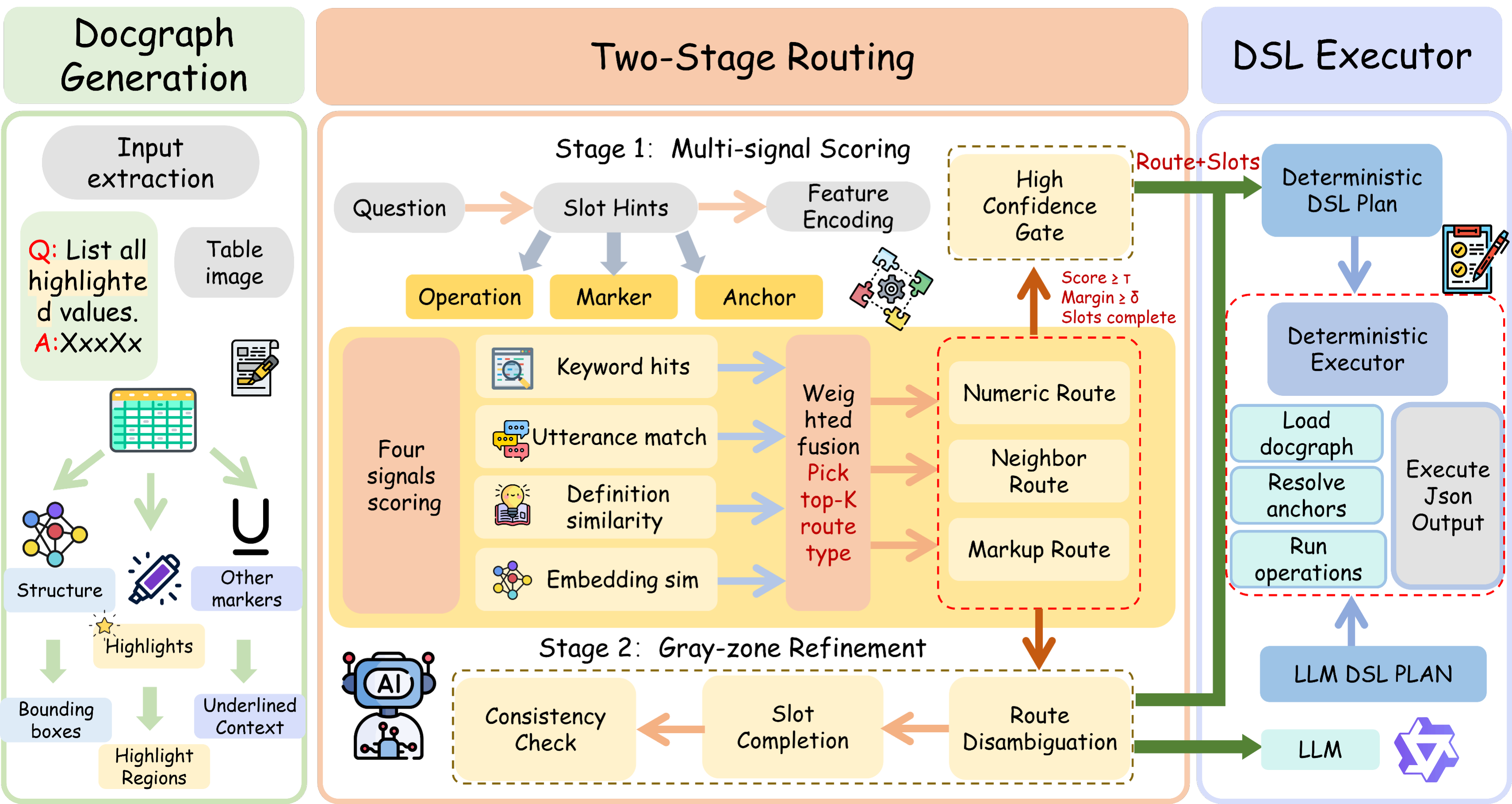}
    \caption{\textbf{Pipeline overview.} The reference pipeline converts the input image into a unified docgraph, uses two-stage routing to determine the solving path, and executes the routed result as a DSL plan on the docgraph to produce the final structured output and its intermediate trace.}
    \label{fig:pipeline_overview}
\end{figure*}

\subsection{Visual Markup Understanding in Text-Dense Documents}

Recent progress in multimodal document understanding has greatly improved reading and reasoning over text-dense images. Representative systems include TextMonkey \cite{liu2026textmonkey}, which strengthens high-resolution visual reading for dense document content, and the mPLUG-DocOwl series \cite{hu2024mplug,hu2025mplug}, which improves document modeling through stronger visual-language alignment and document-oriented instruction tuning. Other document-focused models, including DocPedia \cite{feng2024docpedia}, Vary \cite{wei2024vary}, Marten \cite{wang2025marten}, and DocLayLLM \cite{liao2025doclayllm}, further show that multimodal systems can better handle layout-rich documents, long text regions, and structured extraction. Table structure recognition, especially for complicated layouts, has also been studied as a prerequisite for reliable downstream reasoning \cite{chi2019complicated}. However, these models mainly target general document reading, extraction, and question answering, rather than table reasoning where embedded visual markups themselves carry task-relevant meaning.

A related line of work studies visual prompting and localized interaction. VIP-LLaVA \cite{cai2024vip}, Draw-and-Understand \cite{lin2024draw}, VP-Bench \cite{xu2025vp}, and ControlMLLM \cite{wu2024controlmllm} explore how arrows, points, masks, or free-form visual prompts can steer region-level grounding and localized reasoning. These methods show that explicit visual hints can substantially improve reference resolution. However, in most of these settings, the prompt is externally supplied at inference time. In our setting, by contrast, the cue is already embedded in the table and must be interpreted as part of the original content, requiring the model to infer the cue's semantic role rather than follow an instruction-time hint. This makes markup-conditioned table reasoning qualitatively different from prompt-guided localization.

\subsection{Structured Outputs and Evidence Grounding}

A growing body of work suggests that answer-only evaluation can obscure important failure modes once tasks require structured outputs or explicit evidence. SO-Bench \cite{feng2025so} demonstrates that schema-constrained multimodal outputs expose errors that are invisible under plain answer matching. ExtractBench \cite{ferguson2026extractbench} makes a similar point for structured multimodal extraction by emphasizing format compliance and faithful evidence usage. BBox DocVQA \cite{yu2025bbox} further highlights fine-grained evidence grounding by requiring models to localize supporting regions rather than only producing final answers. Similar diagnostic motivations have also been explored in other structured vision-language settings, such as chart question answering benchmarks that increase diversity and difficulty beyond answer-only scoring \cite{masry2025chartqapro}. Our benchmark follows the same diagnostic spirit, but focuses on a different boundary: whether a model can correctly connect embedded visual markups, table structure, and downstream reasoning steps in a single table image.

While multimodal TableQA and document benchmarks have significantly advanced evaluation for table images and text-dense documents, they typically report end-to-end answer accuracy and do not explicitly separate cue-to-structure association from downstream reasoning behaviors. HighlightBench complements this line by making markup-conditioned evidence selection and subsequent execution directly scorable and diagnosable.
\section{Methodology}
\label{sec:highlightbench}

Our aim is to develop a multimodal benchmark for markup-conditioned reasoning over tables under realistic scientific reading scenarios. HighlightBench is designed as a diagnostic framework that makes markup-conditioned behaviors explicit and automatically scorable, enabling analysis beyond end-to-end answer accuracy and helping identify where a model fails along the perception-to-execution chain.

For models, the central challenge is not merely detecting that a markup exists, but using it as an operational constraint over table structure. This requires correctly binding the cue to its corresponding structural units, interpreting the question under the induced scope, and executing the required table operations.
\subsection{Problem Formulation}
\label{subsec:problem_formulation}

We first formalize each benchmark instance as a tuple \((I, M, Q, A)\), where \(I\) is a table image, \(M\) denotes the visual markups, \(Q\) is the question, and \(A\) is the target answer.\begin{equation}
P(A \mid I, M, Q).
\label{eq:markup_objective}
\end{equation}
Here, \(M\) includes visual cues relevant to reasoning, such as highlighted regions, underlined entries, bold text, arrows, or color-based emphasis. These cues interact with table structure and the question, shaping which evidence should be used and how the reasoning should proceed.

To expose failure sources, we introduce two intermediate variables. Let \(z\) denote the structural evidence indicated by the markups, such as marked cells, headers, rows, columns, or local neighborhoods. Let \(c\) denote the reasoning condition induced by that evidence under the question, such as a filtered subset, a local relation, or a comparison scope. The process can be written as
\begin{equation}
\begin{aligned}
P(A \mid I, M, Q)
&=
\sum_{z,c}
P(A \mid I, c, Q)\,
P(c \mid z, Q)\,
P(z \mid I, M).
\end{aligned}
\label{eq:markup_factorization}
\end{equation}
Intuitively, \(P(z \mid I, M)\) captures whether visual cues are correctly mapped to table structure; \(P(c \mid z, Q)\) captures whether the selected evidence is interpreted in a question-consistent way; and \(P(A \mid I, c, Q)\) captures whether the model can execute the required table reasoning under that condition.

This factorization motivates our task taxonomy and supports coarse-grained attribution of errors to cue-to-structure binding, question-aware interpretation, or downstream execution.

\subsection{Task Families}
\label{subsec:task_families}

HighlightBench decomposes markup-conditioned table reasoning into five complementary task families. Each family emphasizes a different segment of Eq.~\ref{eq:markup_factorization}, allowing targeted evaluation of specific failure modes while still operating on realistic table images.

\begin{figure}[t]
    \centering
    \includegraphics[width=0.9\columnwidth]{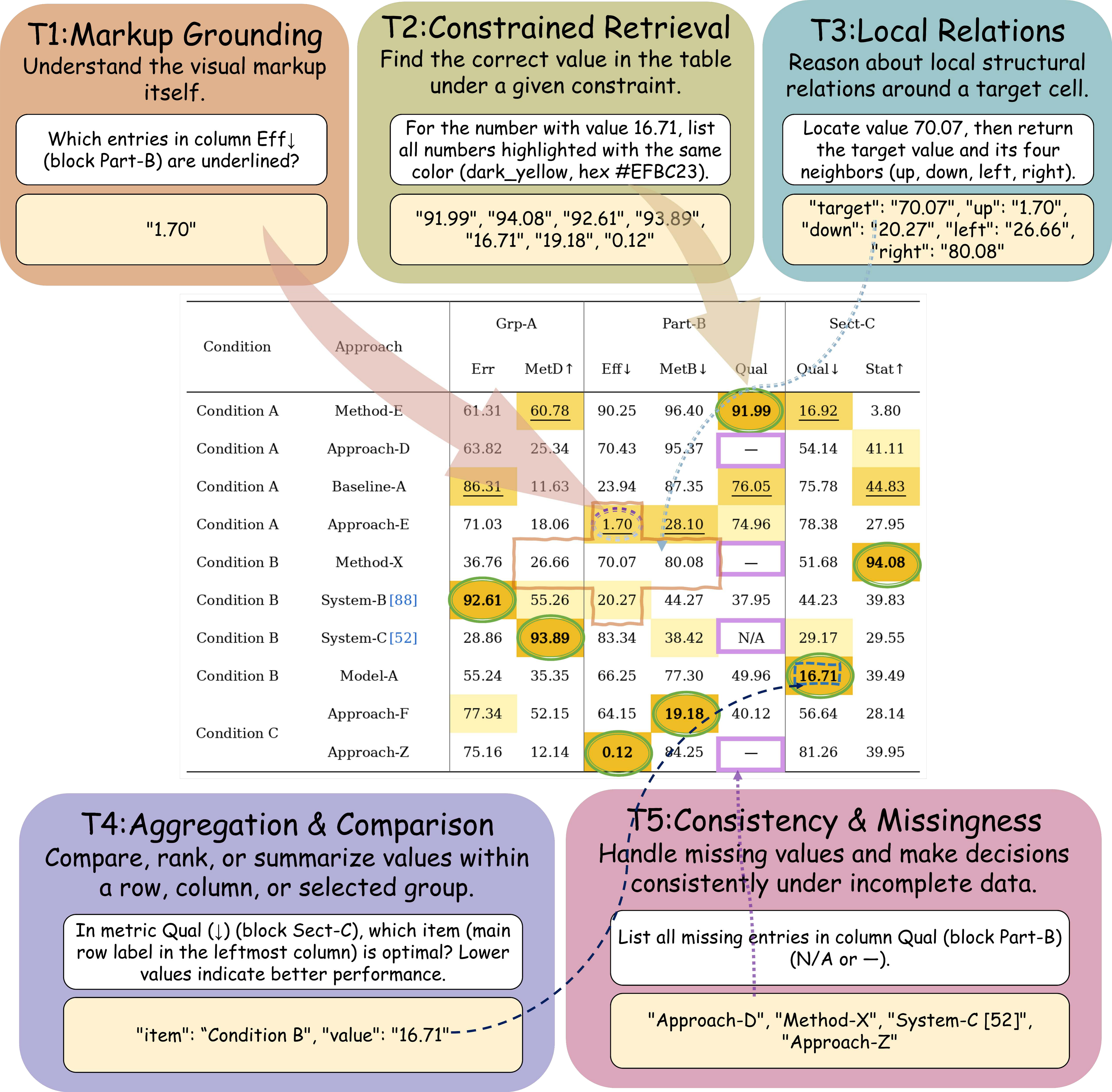}
    \caption{Representative examples of the five task families in HighlightBench. Each card shows the core capability tested by a task family, together with a typical question format and its expected structured output.}
    \label{fig:task_families}
\end{figure}

\textbf{T1. Markup Grounding.}
T1 evaluates cue-to-structure binding: whether highlights, underlines, bold text, and arrows are correctly associated with the intended cells, rows/columns, or headers. It primarily probes the evidence selection term \(P(z \mid I, M)\).

\textbf{T2. Constrained Retrieval.}
T2 evaluates whether a question can be answered by retrieving evidence under markup-conditioned selection, e.g., filtering by a marked subset or using a visual cue as a retrieval anchor. The dominant failure mode is to detect a cue but ignore it during selection, leading to retrieval from an incorrect scope. It primarily probes \(P(c \mid z, Q)\).

\textbf{T3. Local Relations.}
T3 evaluates whether local table topology is preserved, including neighborhood relations along row/column directions and short-range structural dependencies. It bridges evidence localization and downstream reasoning by testing whether models maintain consistent local structure when moving from \(z\) to execution.

\textbf{T4. Aggregation \& Comparison.}
T4 evaluates comparison, ranking, and aggregation operations over subsets determined by the question and markups. It emphasizes faithful execution under the intended scope and mainly probes \(P(A \mid I, c, Q)\).

\textbf{T5. Consistency \& Missingness.}
T5 evaluates reasoning robustness under incomplete, conflicting, or missing evidence (e.g., \texttt{N/A}, dashes, or partially absent fields). It tests whether the model remains consistent with the intended selection and avoids shortcut heuristics when the table contains irregularities, again emphasizing \(P(A \mid I, c, Q)\) under more challenging conditions.

Together, T1--T5 provide complementary diagnostics for the full pipeline from cue binding to retrieval and downstream execution, while maintaining a unified markup-conditioned table reasoning setting.

\subsection{Benchmark Construction}
\label{subsec:benchmark_construction}

HighlightBench contains 446 table images and 3,283 questions in total, including 266 real-world images with 2,268 questions and 180 synthetic images with 1,015 questions. It is constructed from two complementary sources: expert-collected real-world samples and synthetic samples generated through a structured question-driven process. This pairing preserves realistic markup patterns and table irregularities, while enabling controllable and reproducible counterfactual analysis.

\textbf{Real-world samples.}
Human experts select markup-rich tables from CV and NLP papers and write questions for each instance. We verify that each question correctly refers to the markups present in the image and that the reference answer satisfies the corresponding constraints. To improve annotation quality, each instance is independently reviewed by at least two annotators; disagreements are resolved through discussion and final adjudication. We further filter repeated or near-duplicate questions to avoid redundancy. This subset preserves naturally occurring properties of scientific tables, including partial grids, merged cells, multi-level headers, and diverse markup styles.

\textbf{Synthetic samples.}
For the synthetic subset, we start from pre-defined question templates and derive the structural roles required by each instance, including target, distractor, and anchor elements. We then instantiate table contents and visual markups so that the answer is deterministic and automatically scorable. Because this construction process is explicit and controllable, it is fully reproducible and also makes it easy to build counterfactual variants by changing markup conditions while keeping table content fixed. As a quality check, we automatically validate each synthetic instance by executing the underlying generation rule. We also enforce diversity by removing duplicate or near-duplicate generations using template signatures and key entities. This subset improves difficulty control and supports more precise analysis of model failure patterns, while following the same task taxonomy as the real-world subset.

\section{Experiments}
\label{sec:experiments}

HighlightBench is used to evaluate end-to-end QA performance and to examine how multimodal large language models handle visual markups when translating them into reasoning constraints. We aim to understand both overall success rates and the sources of failure: whether they arise from unstable perception of visual cues, from incorrect reasoning with those cues, or from downstream decision errors after the cues have been identified.

This section has two complementary goals. First, we report full-benchmark results to provide an overall performance profile across the five task families. Second, we analyze the benchmark's controlled counterfactual subset, where table content is fixed while only the visual surface changes, allowing us to isolate how markup variations alone affect model behavior. Together, these experiments provide aggregate performance comparisons and direct evidence that HighlightBench can separate basic table reading from markup-conditioned reasoning failures. 

\subsection{Full-Benchmark Evaluation}
\label{subsec:overall_eval}

We evaluate eight representative MLLMs: six open-source models and two closed-source models. Open-source models include Qwen3-VL-8B \cite{bai2025qwen3}, InternVL3.5-8B \cite{wang2025internvl3}, and Qwen2.5-VL-3B \cite{bai2025qwen25vltechnicalreport}. We also evaluate LLaVA1.5-7B \cite{liu2024improved}, MiniCPM-V 2.6 \cite{yao2024minicpm}, and MiniCPM-V 4.5 \cite{yu2025minicpmv45cookingefficient}. Closed-source models are Gemini2.5-Flash \cite{comanici2025gemini} and Claude Sonnet 4 \cite{anthropic2025claude4}. For most model interfaces, we use VLMEvalKit \cite{duan2024vlmevalkit} to standardize model invocation, input formatting, and result collection. All models are evaluated with the same input format, and scores are computed using a post-processed evaluator applied to cleaned JSON outputs.

\begin{table*}[t]
\centering
\caption{Main results on HighlightBench. We report exact-match accuracy (\%) for each task family and the overall score on two subsets: Synthetic and Real-world.}\label{tab:main_full_benchmark}
\small
\setlength{\tabcolsep}{4.2pt}
\renewcommand{\arraystretch}{1.08}
\begin{tabular}{llcccccc}
\toprule
\multirow{2}{*}{\textbf{Model}} 
& \multirow{2}{*}{\textbf{Subset}} 
& \multicolumn{6}{c}{\textbf{Accuracy (\%)}} \\
\cmidrule(lr){3-8}
& & \textbf{T1} & \textbf{T2} & \textbf{T3} & \textbf{T4} & \textbf{T5} & \textbf{All} \\
\midrule

\multirow{2}{*}{Claude Sonnet 4} 
& Synthetic  & 32.7 & \cellcolor{yellow!20}81.9 & 40.0 & 37.5 & 67.1 & 47.6 \\
& Real-world & 56.7 & 59.6 & 42.2 & 60.7 & \cellcolor{yellow!20}68.2 & 58.7 \\
\cmidrule(lr){1-8}

\multirow{2}{*}{Gemini2.5-Flash} 
& Synthetic  & 51.9 & 90.3 & 89.1 & 67.8 & \cellcolor{yellow!20}93.9 & \textbf{73.9} \\
& Real-world & \cellcolor{yellow!20}81.3 & 75.3 & 62.7 & 72.2 & 78.5 & 75.3 \\
\cmidrule(lr){1-8}

\multirow{2}{*}{InternVL3.5-8B} 
& Synthetic  & 29.0 & \cellcolor{yellow!20}81.2 & 17.6 & 24.7 & 71.0 & 40.5 \\
& Real-world & 54.2 & 53.1 & 23.0 & 36.6 & \cellcolor{yellow!20}67.7 & 46.3 \\
\cmidrule(lr){1-8}

\multirow{2}{*}{LLaVA1.5-7B} 
& Synthetic  & \cellcolor{yellow!20}8.1 & 0.7 & 0.0 & 0.0 & 1.0 & 2.7 \\
& Real-world & \cellcolor{yellow!20}3.4 & 1.5 & 1.2 & 0.2 & 0.0 & 1.4 \\
\cmidrule(lr){1-8}

\multirow{2}{*}{MiniCPM-V 2.6} 
& Synthetic  & 17.6 & \cellcolor{yellow!20}59.0 & 7.9 & 13.1 & 14.5 & 20.4 \\
& Real-world & 17.2 & \cellcolor{yellow!20}29.0 & 9.9 & 17.5 & 3.4 & 17.5 \\
\cmidrule(lr){1-8}

\multirow{2}{*}{MiniCPM-V 4.5} 
& Synthetic  & 26.2 & \cellcolor{yellow!20}74.3 & 13.3 & 28.8 & 59.2 & 36.9 \\
& Real-world & \cellcolor{yellow!20}56.2 & 50.0 & 15.5 & 35.9 & 35.7 & 42.8 \\
\cmidrule(lr){1-8}

\multirow{2}{*}{Qwen2.5-VL-3B} 
& Synthetic  & 14.6 & \cellcolor{yellow!20}61.1 & 6.7 & 13.1 & 47.1 & 24.9 \\
& Real-world & 39.4 & 35.6 & 1.2 & 31.3 & \cellcolor{yellow!20}58.9 & 34.7 \\
\cmidrule(lr){1-8}

\multirow{2}{*}{Qwen3-VL-8B} 
& Synthetic  & 37.0 & 74.3 & 18.8 & 31.5 & \cellcolor{yellow!20}74.4 & 44.2 \\
& Real-world & 56.3 & 57.1 & 24.2 & 36.3 & \cellcolor{yellow!20}78.0 & 48.4 \\
\cmidrule(lr){1-8}

\multirow{2}{*}{Ours}
& Synthetic  & 53.4 & \cellcolor{yellow!20}88.9 & 73.3 & 81.5 & 78.9 & 72.3 \\
& Real-world & 72.9 & 74.2 & 70.8 & \cellcolor{yellow!20}82.2 & 79.8 & \textbf{77.1} \\
\bottomrule
\end{tabular}
\end{table*}

\noindent\textbf{Full-benchmark results.}
Table~\ref{tab:main_full_benchmark} reports exact-match accuracy on both the synthetic and real-world subsets. Overall performance varies substantially across models. Among the evaluated end-to-end MLLMs, Gemini2.5-Flash achieves the strongest overall results on both subsets, while most other models show clear degradation on multiple task families. Smaller baselines lag markedly, especially on families that require stable cue handling and multi-step constrained reasoning.

\noindent\textbf{Task-family difficulty is uneven.}
Across models, retrieval-style tasks (T2) are generally the easiest and remain comparatively stable. Two different sources of difficulty emerge beyond T2. Markup grounding (T1) is challenging because it requires reliable cue perception and binding between visual markups and structural units. Local relations (T3) are challenging for a different reason: they require preserving fine-grained table topology and neighborhood structure during reasoning. Building on these requirements, constrained operations (T4) and robustness tests (T5) further amplify upstream errors in cue grounding or local binding, leading to larger performance gaps across models. This uneven profile indicates that HighlightBench separates distinct weaknesses in cue grounding, structural awareness, and constraint-conditioned execution rather than reflecting generic table-reading difficulty.

\noindent\textbf{Reference pipeline as a baseline.}
We also report results from our reference pipeline (\textbf{Ours}) as an additional baseline, and we describe its design choices in Sec.~\ref{sec:pipeline_overview}. Overall, the pipeline performs competitively on both the Synthetic and Real-world subsets and exhibits a distinct profile from end-to-end models. In particular, compared with open-source end-to-end baselines, it tends to perform better on structure- and execution-heavy families (T3--T4), indicating that explicit intermediate structure can stabilize multi-step table operations once the relevant scope is identified. Meanwhile, remaining gaps are primarily linked to the upstream stages that determine which regions should be treated as constraints, which is most evident in cue- and constraint-sensitive families such as T1--T2. This trade-off matches the intended role of the pipeline as a controllable and diagnosable baseline rather than a purely end-to-end predictor.

\subsection{Counterfactual Analysis}
\label{subsec:counterfactual_eval}

To probe how visual markups affect model behavior, we evaluate a controlled counterfactual subset where the table content and question text are fixed and only the visual markup is modified. For each source table, we generate seven image variants: \texttt{n} with no markup; \texttt{Hc}, \texttt{Uc}, and \texttt{Bc}, where highlight, underline, or bold is placed on the ground-truth target of the queried task; and \texttt{Hi}, \texttt{Ui}, and \texttt{Bi}, where the same marker is placed on a distractor. We conduct this analysis on Qwen3-VL-8B.

To make the intervention as controlled as possible for extremum queries, we standardize the comparison direction so that larger values are always preferred, and we ensure that the target is the unique maximum. Under this convention, the congruent variants \texttt{Hc}/\texttt{Uc}/\texttt{Bc} place the marker on the true maximum for the Extremum Reasoning task. This avoids confounding effects from mixed optimization directions and ensures that any observed shift is attributable to markup placement rather than to changes in task semantics.

We focus on five representative sub-tasks, with the specific task definitions listed in Tab.~\ref{tab:exp1_tasks}: Bold Presence, Underline Presence, Highlight Presence, Cell Retrieval, and Extremum Reasoning. Together, these probes reveal several consistent patterns in how the model responds to visual markups under controlled interventions.
\begin{table}[t]
\centering
\caption{Representative sub-tasks used in the counterfactual analysis.}
\label{tab:exp1_tasks}
\small
\setlength{\tabcolsep}{4pt}
\renewcommand{\arraystretch}{1.05}
\begin{tabular}{lll}
\toprule
Task & Family & Brief example \\
\midrule
Bold Presence & T1 & Is a queried cell boldfaced? \\
Underline Presence & T1 & Is a queried cell underlined? \\
Highlight Presence & T1 & Is a queried cell highlighted? \\
Cell Retrieval & T2 & Read the value at a specified row/column. \\
Extremum Reasoning & T4 & Return the item/value with the column maximum. \\
\bottomrule
\end{tabular}
\end{table}

\begin{figure}[tb]
\centering
\includegraphics[width=0.6\linewidth]{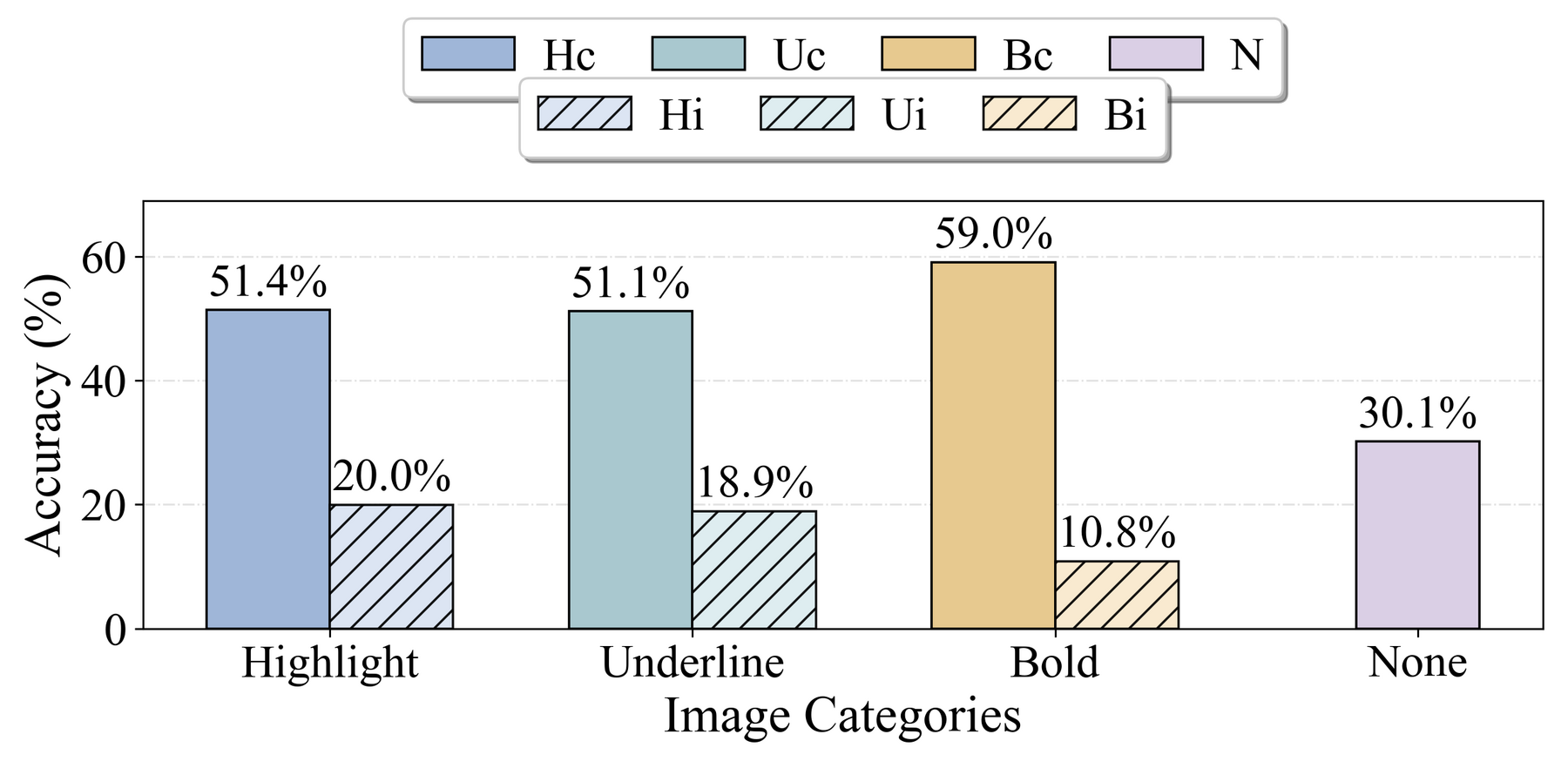}
\caption{Extremum Reasoning accuracy under the counterfactual variants.}
\label{fig:4_1}
\end{figure}

\begin{figure}[tb]
\centering
\begin{subfigure}{0.47\linewidth}
\centering
\includegraphics[width=\linewidth]{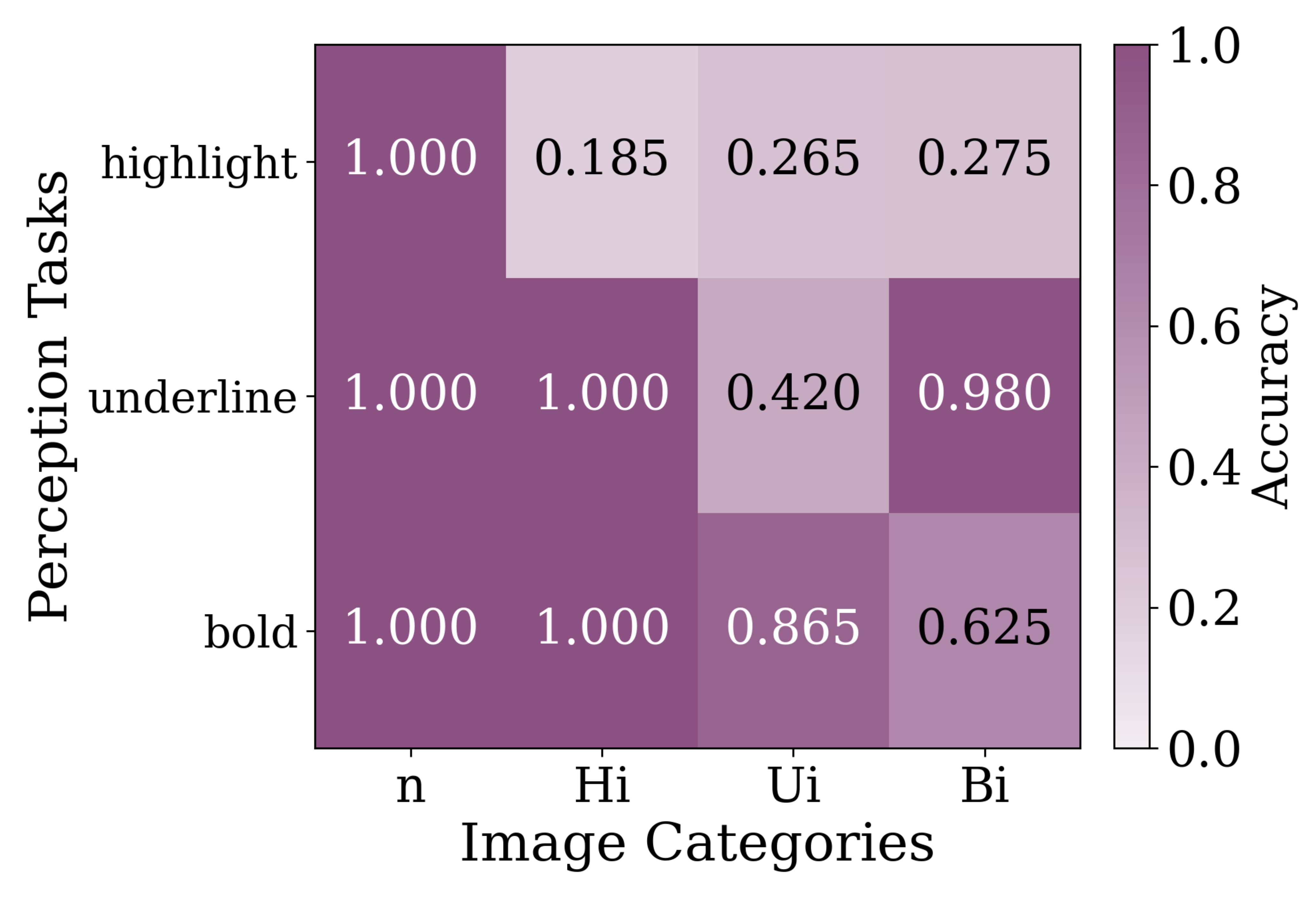}
\caption{Incongruent: \texttt{n}, \texttt{Hi}, \texttt{Ui}, \texttt{Bi}}
\label{fig:4_2}
\end{subfigure}
\hfill
\begin{subfigure}{0.47\linewidth}
\centering
\includegraphics[width=\linewidth]{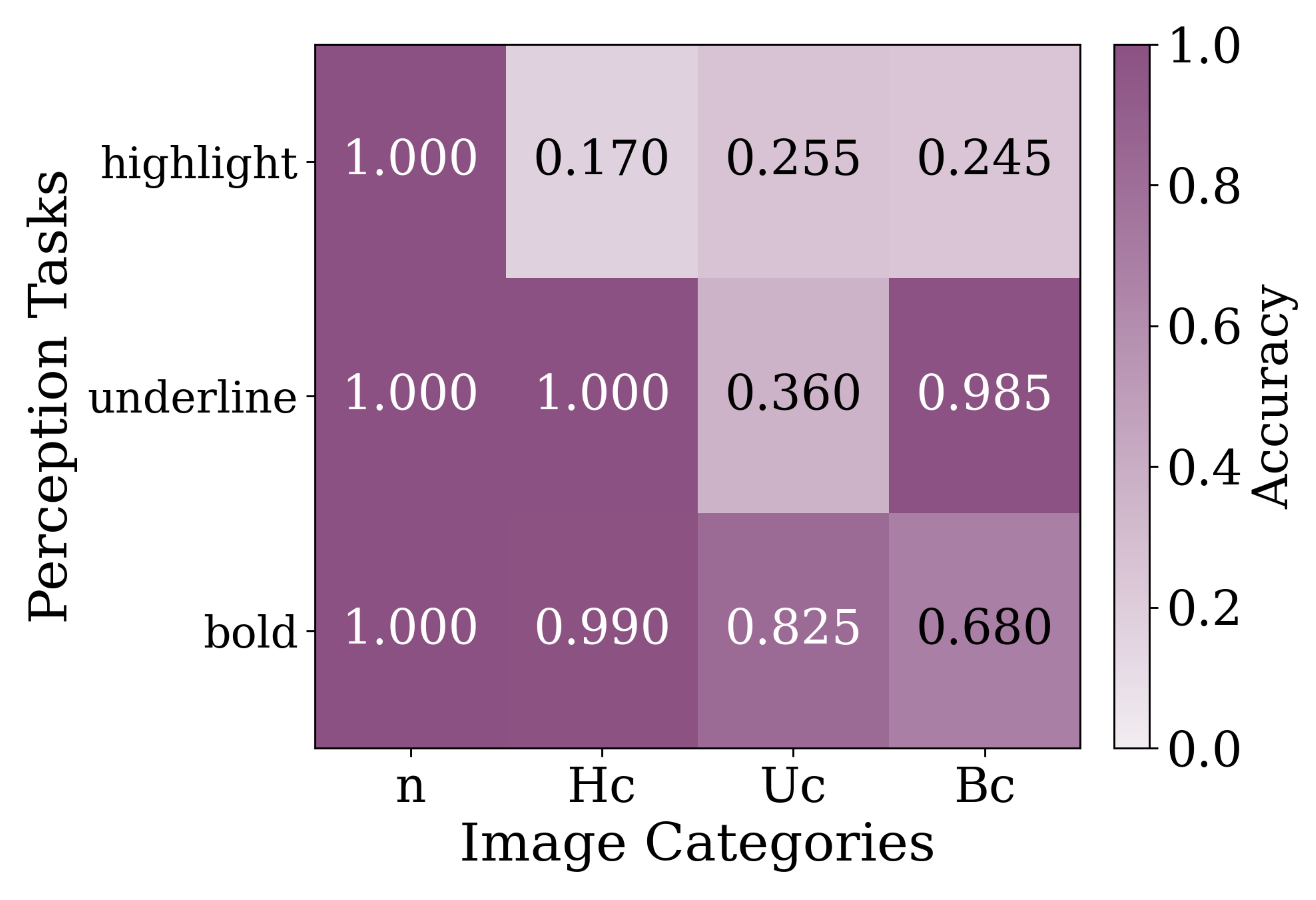}
\caption{Congruent: \texttt{n}, \texttt{Hc}, \texttt{Uc}, \texttt{Bc}}
\label{fig:4_3}
\end{subfigure}
\caption{Perception-task accuracy under counterfactual variants. Rows correspond to Bold Presence, Underline Presence, and Highlight Presence.}
\label{fig:4_2_4_3}
\end{figure}

\begin{figure*}[tb]
\centering
\begin{subfigure}[t]{0.42\textwidth}
\centering
\includegraphics[width=0.92\linewidth]{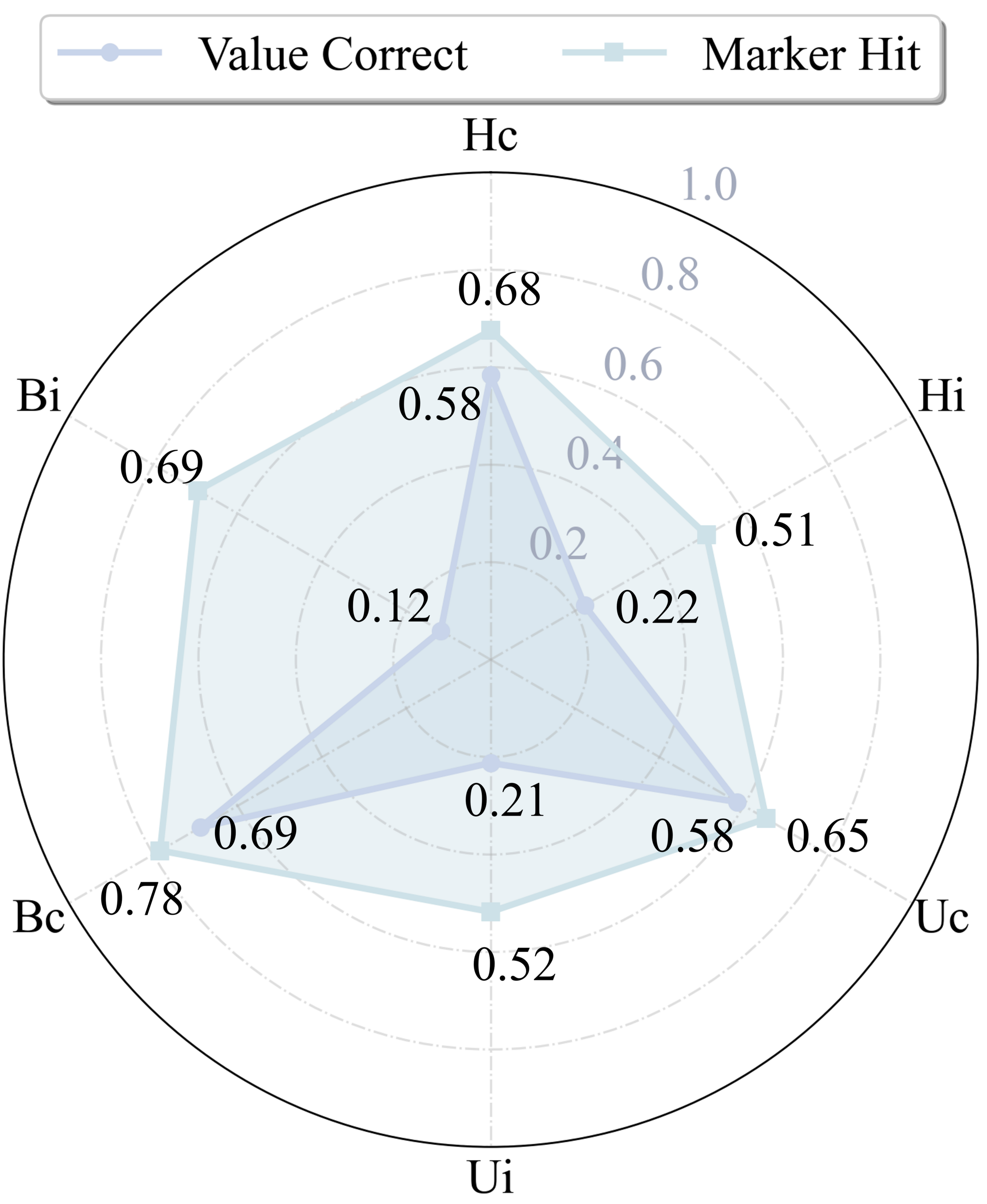}
\caption{Marker-hit vs. value-correct rates. Marker-hit checks whether the predicted value lands on a marked cell; value-correct checks whether the predicted value matches the ground-truth answer.}
\label{fig:4_5}
\end{subfigure}
\hfill
\begin{subfigure}[t]{0.42\textwidth}
\centering
\includegraphics[width=0.92\linewidth]{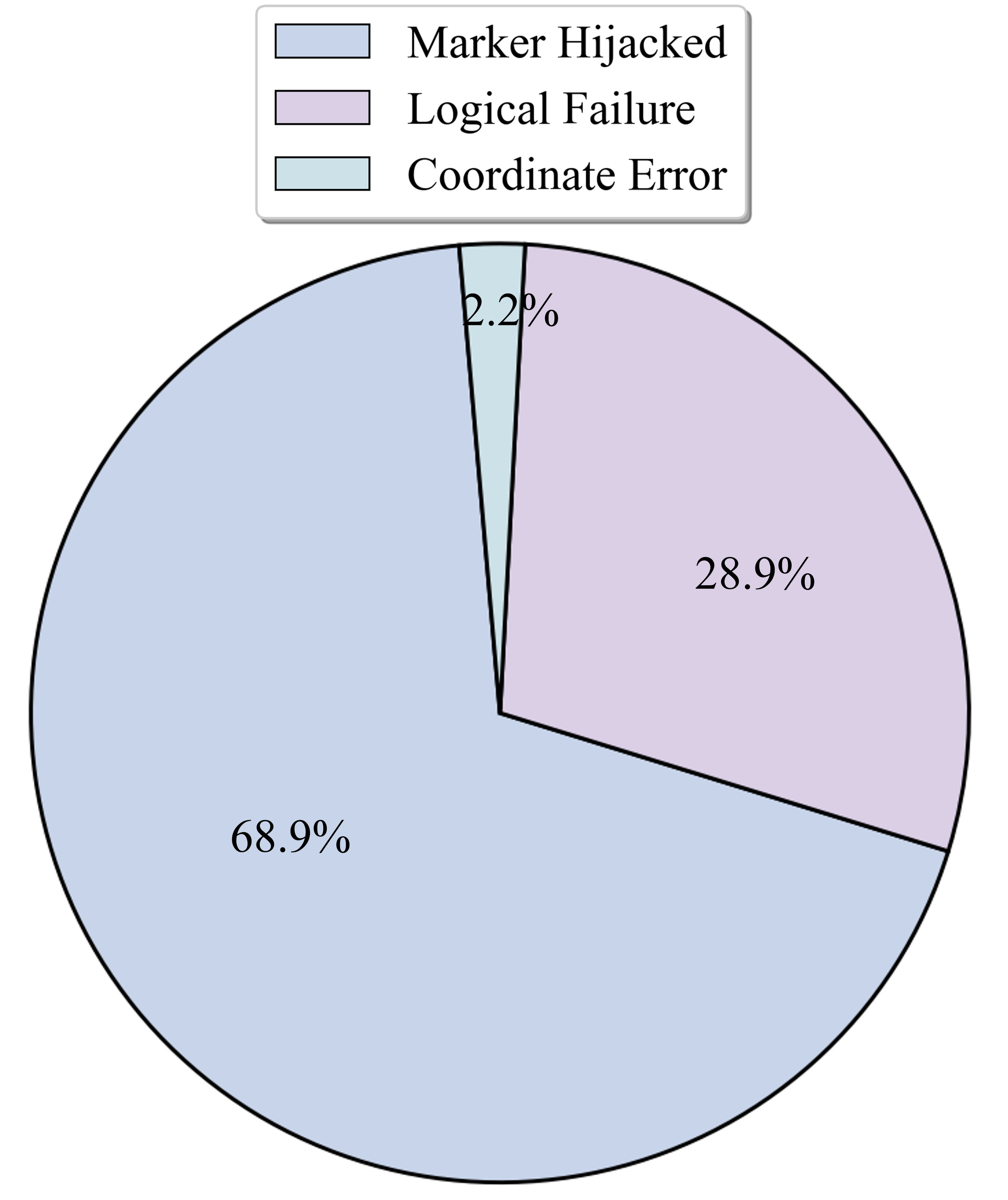}
\caption{Failure attribution under counterfactual variants.}
\label{fig:4_6}
\end{subfigure}
\caption{Mechanism probes under counterfactual variants. (a) Marker-hit rate versus value-correct rate across variants. (b) Error attribution of failed cases based on observable outcomes.}
\label{fig:4_5_4_6}
\end{figure*}

\noindent\textbf{Markers materially change the final decision.}
The clearest effect appears in Extremum Reasoning. As shown in Fig.~\ref{fig:4_1}, changing only the marker placement produces large performance shifts: congruent settings consistently improve performance over the no-markup baseline, while incongruent settings consistently reduce it. Because table content and question text are identical across variants, these shifts are driven by the markup intervention itself, showing that visual markers can directly steer the final selection.

\noindent\textbf{The main issue is not basic reading.}
This becomes clearer when compared with the Cell Retrieval control. Under the same counterfactual protocol and evaluation run, Cell Retrieval remains consistently high across the seven variants, staying within 94.5\%--96.5\%. This stability indicates that the large swings observed in T1 and T4 are unlikely to be caused by generic table-reading failures. Instead, the main difference lies in how the model reacts to the added markers once they are present.

\noindent\textbf{Explicit recognition shows a negative bias and highlight is harder.}
A different pattern appears in the perception tasks (Fig.~\ref{fig:4_2_4_3}). For bold and underline, accuracy is very high in variants where the queried markup is absent, suggesting that the model often answers ``not present'' correctly and may default to negative responses. In contrast, highlight presence is substantially less stable across variants, indicating that explicit recognition of highlight is more difficult and more prone to confusion. This gap helps explain why a model can be influenced by markups in downstream decisions without reliably identifying those markups at the explicit reporting level.

\noindent\textbf{Marked distractors dominate many failures.}
Fig.~\ref{fig:4_5_4_6}\subref{fig:4_5} shows a consistent gap between marker-hit rate and value-correct rate, indicating that predictions often land on marked values even when they are incorrect. Using value correctness and marker hit as observable signals, we further group failed cases into three attribution categories in Fig.~\ref{fig:4_5_4_6}\subref{fig:4_6}. Marker Hijacked refers to errors where the predicted value is incorrect but hits a marked cell, suggesting that the marker pulls the decision toward a distractor. Logical Failure refers to incorrect values without marker hit, indicating that the error is not directly driven by the marked region. Coordinate Error refers to cases where the value is correct but the associated item is incorrect, suggesting a binding or localization mismatch. The breakdown shows that most failures fall into Marker Hijacked, with smaller portions explained by Logical Failure and Coordinate Error. This attribution is operational rather than mechanistic, but it provides concrete evidence that marked distractors are a dominant error source under conflicting visual cues.

Taken together, these results show that markups can systematically bias decisions under otherwise fixed evidence, motivating a more explicit inference interface in Sec.~\ref{sec:pipeline_overview}.

\section{Pipeline Overview}
\label{sec:pipeline_overview}

Building on the counterfactual findings in Sec.~\ref{subsec:counterfactual_eval}, we introduce a reference pipeline as an explicit interface for markup-conditioned reasoning. Rather than replacing end-to-end MLLMs, the pipeline externalizes intermediate decisions that are otherwise hidden in a single response, making the solving process inspectable and enabling finer-grained failure attribution.

As shown in Fig.~\ref{fig:pipeline_overview}, the pipeline decomposes solving into three stages. It first converts the input into a structured docgraph that explicitly represents table content, table topology, and detected markups. It then selects a solving route and resolves task-specific constraints under a two-stage routing scheme. Finally, it executes the resulting plan with a deterministic DSL executor and returns both the final answer and intermediate traces in a unified schema.

\textbf{Docgraph Generation.}
The first design choice is to separate the visual surface from the evidence used for reasoning. The docgraph stage converts the input image into an explicit intermediate representation that organizes textual content, table structure, and markup signals in a shared space. Markups are stored as structured attributes bound to specific cells or regions, rather than remaining as an implicit cue in a free-form response. This representation encourages later stages to operate on explicit units and relations, reduces reliance on raw visual salience, and makes binding mistakes observable.

\textbf{Two-Stage Routing.}
A second design choice is to separate route commitment from constraint completion. The first routing stage performs coarse route selection to determine the route type and required operators based on the question form and readily available signals from the parsed context. The second stage is reserved for uncertainty resolution. It revisits cue-related conditions and completes missing constraints when the initial route is underspecified or internally inconsistent, for example when a task requires a markup-conditioned subset but the extracted cue evidence is weak or ambiguous. Compared with a single-stage router, this staged design reduces the chance that early interpretation errors are directly propagated into execution and makes failures easier to attribute to route selection versus constraint completion.

\textbf{DSL Execution.}
The final design choice is to enforce constraint-faithful computation with deterministic execution. The routed outcome is converted into an executable DSL plan and executed on the docgraph by a deterministic executor. Constraint-conditioned operations such as filtering, extremum selection, aggregation, and consistency checks are implemented as explicit operators with fixed semantics. This reduces free-form variability, makes results reproducible, and produces intermediate traces that expose whether a failure arises from cue-to-structure binding, question-aware interpretation, or downstream execution.

Overall, the reference pipeline is a problem-driven interface for markup-conditioned reasoning that emphasizes explicit representations, staged interpretation, and traceable execution. By making intermediate decisions inspectable, it serves both as a reproducible baseline and as a diagnostic tool for understanding and mitigating failures in current multimodal models.

\section{Conclusion}

We present \textbf{HighlightBench}, a diagnostic benchmark for markup-conditioned reasoning over scientific tables. The benchmark covers five complementary task families and includes a controlled counterfactual subset where only markup placement varies under fixed table content and question text.

Across both full-benchmark evaluation and counterfactual analysis, we find that current MLLMs do not consistently treat visual markups as executable reasoning constraints. Models can be relatively strong at direct retrieval, yet remain fragile when correct cue grounding and constraint-conditioned execution must be combined. The counterfactual results show that marker placement alone can systematically shift decisions, revealing a failure mode where visually salient cues can dominate outcomes even when the underlying evidence is unchanged.

Looking forward, these results suggest that progress in table understanding should be assessed and improved through diagnostic decomposition rather than answer-only scoring. Evaluations should separate cue grounding, constraint formation, and constrained execution, and systems should expose intermediate decisions to make errors traceable and correctable. More broadly, treating markups as first-class operational constraints is important for building multimodal tools that can reliably support scientific reading, verification, and table-centric document intelligence.

\bibliographystyle{splncs04}
\bibliography{main}

\clearpage
\appendix
\renewcommand{\theHsection}{appendix.\Alph{section}}
\renewcommand{\theHsubsection}{appendix.\Alph{section}.\arabic{subsection}}
\begin{center}
{\Large\bfseries Appendix}
\end{center}
\vspace{0.6em}
\section{Qualitative Failure Cases}
\label{app:dataset_case_studies}

To complement the quantitative results in the main paper, we present five qualitative cases from HighlightBench. The benchmark is organized into five task families and further decomposed into 21 fine-grained subtasks. We select one subtask from each family and show a representative error under a shared visual context. Each case box reports the subtask name, the original question, and the ground-truth and predicted outputs; the accompanying paragraph summarizes the failure in terms of the perception-to-execution chain.

\begin{figure}[!h]
\centering
\IfFileExists{figures/0070.png}{%
  \includegraphics[width=0.86\linewidth]{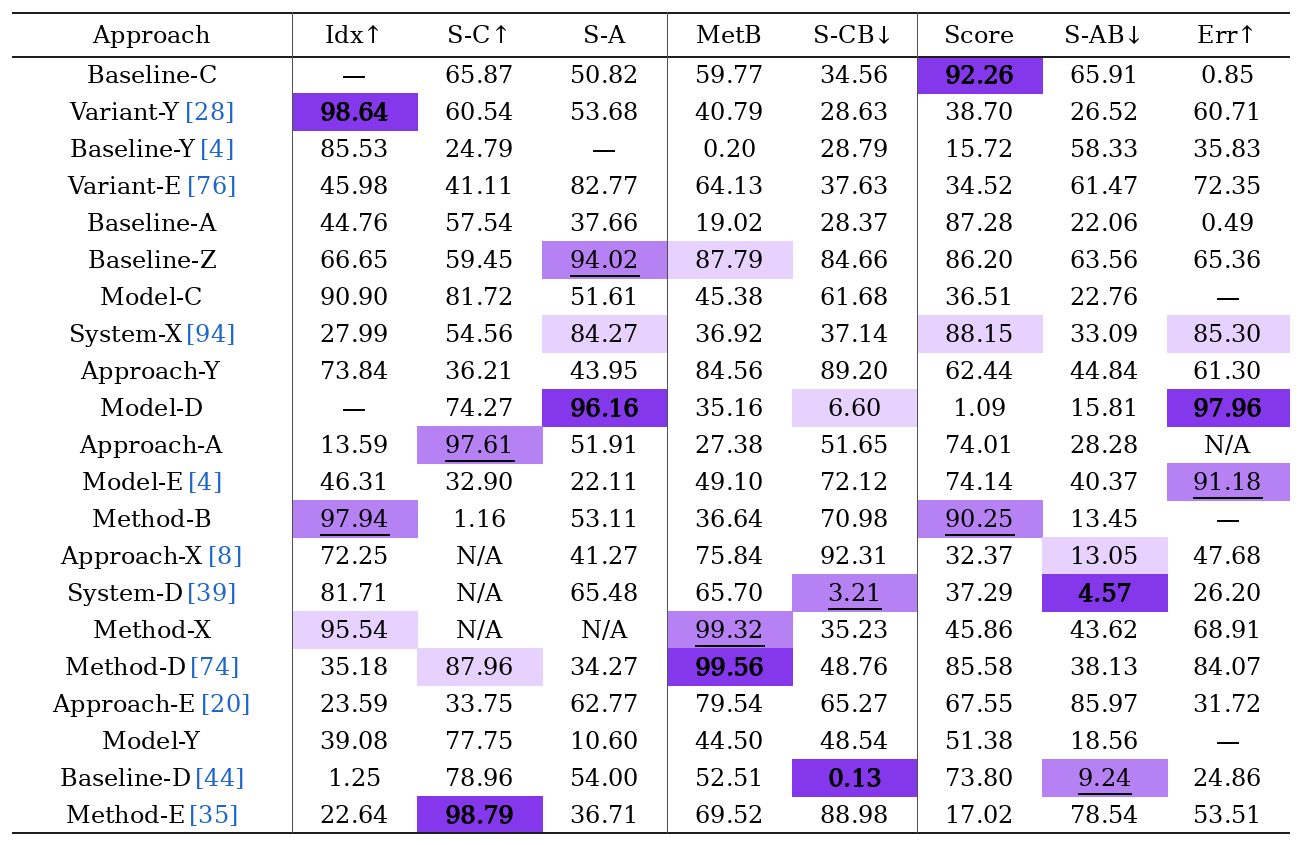}%
}{%
  \fbox{\parbox[c][0.24\textheight][c]{0.85\linewidth}{\centering
  Missing shared image: \texttt{figures/0070.png}}}%
}
\caption{Shared table image for the qualitative failure cases.}
\label{fig:shared_case_image}
\end{figure}
\FloatBarrier

\begin{casebox}{A.1 Markup Grounding}
\small
\setlength{\parskip}{0.3em}

\textbf{Task Name.} \texttt{Marked Value Listing}

\textbf{Question.}
List numbers highlighted in mid\_purple (hex \#B681F3). Output only a JSON
array of strings. Example: \texttt{["12.34","56.78"]}.

\begin{tabularx}{\linewidth}{@{}>{\raggedright\arraybackslash}p{0.28\linewidth}X@{}}
\gtlabel &
\gtcell{\footnotesize\ttfamily ["94.02", "97.61", "91.18", "97.94", "90.25", "3.21", "99.32", "9.24"]} \\[0.55em]
\predlabel &
\predcell{\footnotesize\ttfamily ["98.64", "97.94", "97.94", "97.94", "97.94", "97.94", "97.94", "97.94", ...]}
\end{tabularx}
\end{casebox}

\noindent
This case shows that the model does not produce a bounded set of marked values: the ground-truth output contains eight highlighted entries, whereas the prediction degenerates into repeated values. The failure is consistent with unstable cue-to-structure binding that prevents a reliable marked set from being formed.

\begin{casebox}{A.2 Constrained Retrieval}
\small
\setlength{\parskip}{0.3em}

\textbf{Task Name.} \texttt{Value Filtering}

\textbf{Question.}
In column S-CB$\downarrow$, list all items with value $< 48.54$. Return a JSON
array of \texttt{\{"item":"...","value":"..."\}} objects. Exclude
\texttt{N/A} and \texttt{—}.

\begin{tabularx}{\linewidth}{@{}>{\raggedright\arraybackslash}p{0.28\linewidth}X@{}}
\gtlabel &
\gtcell{\footnotesize\ttfamily [{"item":"Baseline-C","value":"34.56"}, {"item":"Variant-Y [28]","value":"28.63"}, {"item":"Baseline-Y [4]","value":"28.79"}, {"item":"Variant-E [76]","value":"37.63"}, {"item":"Baseline-A","value":"28.37"}, {"item":"System-X [94]","value":"37.14"}, {"item":"Model-D","value":"6.60"}, {"item":"System-D [39]","value":"3.21"}, {"item":"Method-X","value":"35.23"}, {"item":"Baseline-D [44]","value":"0.13"}]} \\[0.55em]
\predlabel &
\predcell{\footnotesize\ttfamily [{"item":"Baseline-C","value":"34.56"}, {"item":"Variant-Y [28]","value":"28.63"}, {"item":"Baseline-Y [4]","value":"28.79"}, {"item":"Variant-E [76]","value":"37.63"}, {"item":"Baseline-A","value":"28.37"}, {"item":"Baseline-Z","value":"84.66"}, {"item":"Model-C","value":"61.68"}, {"item":"System-X [94]","value":"37.14"}, {"item":"Approach-Y","value":"89.20"}, {"item":"Model-D","value":"6.60"}, {"item":"Approach-A","value":"51.65"}, {"item":"Model-E [4]","value":"72.12"}, {"item":"Method-B","value":"70.98"}, ...]}
\end{tabularx}
\end{casebox}

\noindent
This case shows that correct column access is insufficient for constrained retrieval: the prediction includes several items below the threshold but also retains multiple entries above it. The pattern indicates that the inequality constraint is not consistently preserved during selection.

\begin{casebox}{A.3 Local Relations}
\small
\setlength{\parskip}{0.3em}

\textbf{Task Name.} \texttt{Neighbor Indexing}

\textbf{Question.}
For the DG cell at \((r=7,c=7)\), return target/up/down/left/right as JSON
\texttt{\{"target":"12.34", "up":"11.11", "down":"56.78", "left":"N/A", "right":"90.12"\}}.
DG includes numeric/data cells only (no headers; no left header columns). If a
neighbor cell contains \texttt{N/A} or \texttt{—}, return that token string.

\begin{tabularx}{\linewidth}{@{}>{\raggedright\arraybackslash}p{0.28\linewidth}X@{}}
\gtlabel &
\gtcell{\footnotesize\ttfamily \{"target":"22.76", "up":"63.56", "down":"33.09", "left":"36.51", "right":"—"\}} \\[0.55em]
\predlabel &
\predcell{\footnotesize\ttfamily \{"target":"90.12", "up":"97.96", "down":"91.18", "left":"N/A", "right":"90.25"\}}
\end{tabularx}
\end{casebox}

\noindent
This case shows a global anchoring error: the returned neighborhood fields form a locally coherent pattern but are centered on the wrong table region. This is consistent with a failure to bind the queried coordinate to the intended structural anchor prior to execution.

\begin{casebox}{A.4 Aggregation \& Comparison}
\small
\setlength{\parskip}{0.3em}

\textbf{Task Name.} \texttt{Rank Selection}

\textbf{Question.}
Return the 3 highest rows in S-A in descending order as a JSON array of
\texttt{\{"item":"...","value":"..."\}} objects. Exclude \texttt{N/A} and
\texttt{—}.

\begin{tabularx}{\linewidth}{@{}>{\raggedright\arraybackslash}p{0.28\linewidth}X@{}}
\gtlabel &
\gtcell{\footnotesize\ttfamily [{"item":"Model-D","value":"96.16"}, {"item":"Baseline-Z","value":"94.02"}, {"item":"System-X [94]","value":"84.27"}]} \\[0.55em]
\predlabel &
\predcell{\footnotesize\ttfamily [{"item":"Model-C","value":"51.61"}, {"item":"Model-D","value":"96.16"}, {"item":"Baseline-Z","value":"94.02"}]}
\end{tabularx}
\end{casebox}

\noindent
This case shows that ranking is sensitive to small candidate-set corruption: the model identifies the top two entries but replaces the true third item with an incorrect candidate. The error suggests that downstream aggregation/comparison amplifies earlier selection noise rather than correcting it.

\begin{casebox}{A.5 Consistency \& Missingness}
\small
\setlength{\parskip}{0.3em}

\textbf{Task Name.} \texttt{Column Missingness}

\textbf{Question.}
In Err$\uparrow$, which rows have missing values (\texttt{N/A} or
\texttt{—})? Return a JSON array of row names. Example:
\texttt{["System-A","System-B"]}.

\begin{tabularx}{\linewidth}{@{}>{\raggedright\arraybackslash}p{0.28\linewidth}X@{}}
\gtlabel &
\gtcell{\footnotesize\ttfamily ["Model-C", "Approach-A", "Method-B", "Model-Y"]} \\[0.55em]
\predlabel &
\predcell{\footnotesize\ttfamily ["Model-D", "Approach-A"]}
\end{tabularx}
\end{casebox}

\noindent
This case shows inconsistent set tracking under missing evidence: the prediction omits valid missing rows and introduces an unsupported one. The behavior indicates that the model does not maintain a conservative, evidence-faithful missingness set.

\noindent
Taken together, these cases reinforce the main diagnostic claim of HighlightBench: failures can emerge at different points along the perception-to-execution chain rather than from a single generic weakness in table reading.

\section{Per-Subtask Results}
\label{app:per_subtask_results}

\begin{figure}[!h]
\centering
\IfFileExists{figures/heatmap.png}{%
  \includegraphics[width=\linewidth]{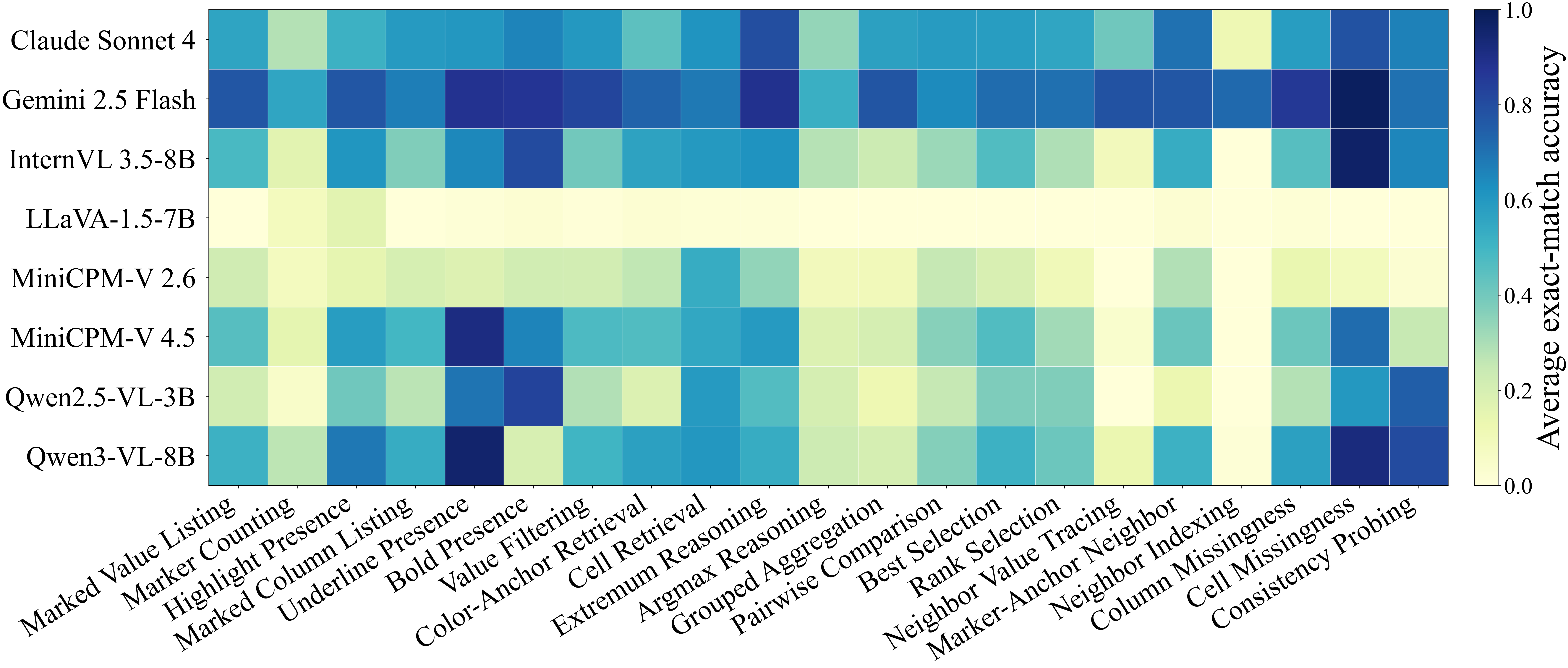}%
}{%
  \fbox{\parbox[c][0.22\textheight][c]{0.95\linewidth}{\centering
  Missing heatmap image: \texttt{figures/heatmap.png}}}%
}
\caption{Per-subtask performance heatmap across models on HighlightBench. Rows denote models and columns denote the 21 subtasks. Color indicates average exact-match accuracy on the full benchmark.}
\label{fig:appB_subtask_heatmap}
\end{figure}

Figure~\ref{fig:appB_subtask_heatmap} reports per-subtask exact-match accuracy for eight representative models on the full benchmark, aggregating over all tables and questions. The heatmap provides a compact view of how performance is distributed across the 21 subtasks, complementing family-level results with a finer diagnostic breakdown.

The columns exhibit clear heterogeneity. Some subtasks are handled reasonably well by most models, while others remain challenging across the board. This is consistent with the benchmark design. Task families group subtasks by the type of markup-conditioned skill being tested rather than by difficulty. Subtasks within the same family can therefore differ in the requirements they impose, including cue binding, schema-constrained outputs, and scope preservation across multiple steps. The heatmap also reveals model-specific profiles. Strengths concentrate on a subset of subtasks and persistent weaknesses remain on a small set of stress-test columns, which offers a focused target set for follow-up diagnosis.

\FloatBarrier
\section{Notation}
\label{app:notation}

Symbols used in the problem formulation and task taxonomy are summarized below. The main paper introduces markup-conditioned TableQA through the tuple $(I,M,Q,A)$ and the intermediate variables $(z,c)$ for diagnostic factorization. The reference pipeline further exposes routing and planning variables to make intermediate decisions inspectable. We list the symbols and the 21 subtasks for quick reference.
\subsection{Symbols}
\label{app:notation_symbols}

\setlength{\LTpre}{6pt}
\setlength{\LTpost}{6pt}

\small
\setlength{\tabcolsep}{5pt}
\renewcommand{\arraystretch}{1.15}
\begin{longtable}{@{}p{0.18\linewidth}p{0.78\linewidth}@{}}
\caption{Notation used in HighlightBench and the reference pipeline.}
\label{tab:notation}\\
\toprule
\textbf{Symbol} & \textbf{Meaning} \\
\midrule
\endfirsthead
\toprule
\textbf{Symbol} & \textbf{Meaning} \\
\midrule
\endhead
\bottomrule
\endfoot
\bottomrule
\endlastfoot

$I$ & Table image. \\
$M$ & Visual markups in $I$ that are relevant to reasoning, including highlights, underlines, bold text, arrows, and color-based emphasis. \\
$Q$ & Question associated with $I$. \\
$A$ & Target structured answer under the task-specific output schema. \\
\midrule
$z$ & Structural evidence indicated by $M$, such as marked cells, headers, rows, columns, or local neighborhoods. \\
$c$ & Reasoning condition induced from $(z,Q)$, such as a filtered subset, a local relation, or a comparison scope. \\
\midrule
$P(A \mid I, M, Q)$ & Markup-conditioned TableQA objective. \\
$P(z \mid I, M)$ & Cue-to-structure binding that maps markups to structural units. \\
$P(c \mid z, Q)$ & Question-aware interpretation that converts evidence into an executable condition. \\
$P(A \mid I, c, Q)$ & Downstream execution under condition $c$. \\
\midrule
$\mathcal{G}$ & Docgraph, a structured intermediate representation constructed from $I$ and used for execution. \\
$r$ & Route type selected by the two-stage router, indicating the solving path. \\
$\mathbf{s}$ & Slot assignment associated with the selected route, specifying required fields such as operation, marker type, or anchor. \\
$\rho$ & Router confidence score used for gating high-confidence decisions versus gray-zone refinement. \\
$\tau$ & Router score threshold used by the high-confidence gate. \\
$\delta$ & Router margin threshold used by the high-confidence gate. \\
$m$ & Routing margin used by the high-confidence gate, computed for the stage-1 route ranking. \\
$\kappa$ & Slot-completeness indicator for the selected route, where $\kappa=\textbf{true}$ means required slots are filled. \\
$p$ & DSL plan executed on $\mathcal{G}$, produced deterministically or by an LLM planner. \\
$\xi$ & Intermediate trace recording route, slots, plan, and execution outputs. \\
\midrule
$T_1 \dots T_5$ & Task families. $T_1$ Markup Grounding, $T_2$ Constrained Retrieval, $T_3$ Local Relations, $T_4$ Aggregation \& Comparison, $T_5$ Consistency \& Missingness. \\
\end{longtable}
\normalsize

\subsection{Subtasks}
\label{app:notation_subtasks}

\small
\setlength{\tabcolsep}{5pt}
\renewcommand{\arraystretch}{1.12}
\begin{longtable}{@{}p{0.24\linewidth}p{0.72\linewidth}@{}}
\caption{List of the 21 subtasks grouped by the five task families.}
\label{tab:subtask_list}\\
\toprule
\textbf{Family} & \textbf{Subtasks} \\
\midrule
\endfirsthead
\toprule
\textbf{Family} & \textbf{Subtasks} \\
\midrule
\endhead
\bottomrule
\endfoot
\bottomrule
\endlastfoot
$T_1$ Markup Grounding &
Marked Value Listing, Marker Counting, Highlight Presence, Marked Column Listing, Underline Presence, Bold Presence \\
\midrule
$T_2$ Constrained Retrieval &
Value Filtering, Color-Anchor Retrieval, Cell Retrieval \\
\midrule
$T_3$ Local Relations &
Neighbor Value Tracing, Marker-Anchor Neighbor, Neighbor Indexing \\
\midrule
$T_4$ Aggregation \& Comparison &
Extremum Reasoning, Argmax Reasoning, Grouped Aggregation, Pairwise Comparison, Best Selection, Rank Selection \\
\midrule
$T_5$ Consistency \& Missingness &
Column Missingness, Cell Missingness, Consistency Probing \\
\end{longtable}
\normalsize

\section{Pseudocode}
\label{app:pseudocode}
\setcounter{algorithm}{0}

High-level pseudocode is provided for the reference pipeline used to produce reproducible baselines and diagnostic traces. The pseudocode abstracts away implementation details such as specific model backends, while preserving the interfaces between major stages. The router outputs a route type $r$ and a slot assignment $\mathbf{s}$, and the pipeline records an intermediate trace $\xi$ for diagnosis.

\begin{algorithm}[H]
\caption{Batch evaluation protocol.}
\label{alg:pipeline_eval}
\small
\begin{algorithmic}[1]
\REQUIRE Dataset $\mathcal{D}=\{(I_i,Q_i,A_i)\}_{i=1}^N$
\ENSURE Aggregate metrics $\mathcal{M}$

\FOR{$i \leftarrow 1$ \TO $N$}
    \STATE (\textit{ans}$_i$, $\xi_i$) $\leftarrow$ \textsc{PipelineInference}$(I_i,Q_i)$
\ENDFOR
\STATE $\mathcal{M} \leftarrow \textsc{Score}(\{\textit{ans}_i\}_{i=1}^N,\{A_i\}_{i=1}^N)$
\RETURN $\mathcal{M}$
\end{algorithmic}
\end{algorithm}

\begin{algorithm}[H]
\caption{Reference pipeline inference.}
\label{alg:pipeline_inference}
\small
\begin{algorithmic}[1]
\REQUIRE Example $(I,Q)$
\REQUIRE Gate thresholds $\tau$ (score) and $\delta$ (margin)
\ENSURE Final answer \textit{ans} and trace $\xi$

\STATE $\mathcal{G} \leftarrow f_{\textsc{dg}}(I)$

\STATE $(r_1,\mathbf{s}_1,\rho_1,m_1,\kappa_1) \leftarrow f_{\textsc{route}}^{(1)}(Q,\mathcal{G})$ \COMMENT{Stage 1 routing}

\IF{$\rho_1 \ge \tau \ \wedge\  m_1 \ge \delta \ \wedge\  \kappa_1$}
    \STATE $(r,\mathbf{s},\rho) \leftarrow (r_1,\mathbf{s}_1,\rho_1)$
\ELSE
    \STATE $(r,\mathbf{s},\rho) \leftarrow f_{\textsc{route}}^{(2)}(Q,\mathcal{G},r_1,\mathbf{s}_1)$ \COMMENT{Stage 2 refinement}
\ENDIF

\STATE $p \leftarrow f_{\textsc{plan}}(Q,\mathcal{G},r,\mathbf{s})$
\STATE $p \leftarrow f_{\textsc{stb}}(p,Q,\mathcal{G},r,\mathbf{s})$
\STATE $y \leftarrow f_{\textsc{exec}}(p,\mathcal{G})$
\STATE $y \leftarrow f_{\textsc{rec}}(y,Q,\mathcal{G},r,\mathbf{s})$
\STATE \textit{ans} $\leftarrow f_{\textsc{post}}(y,Q,r)$

\STATE $\xi \leftarrow \{\mathcal{G}, r, \mathbf{s}, \rho, p, y\}$
\RETURN (\textit{ans}, $\xi$)
\end{algorithmic}
\end{algorithm}

\end{document}